\pdfoutput=1
\documentclass[a4paper,11pt]{article}

\usepackage{fourier}
\usepackage{color}
 \usepackage{graphicx}
\usepackage{url}
\usepackage[affil-it]{authblk}
\usepackage{amsmath}
\usepackage{amsfonts}
\usepackage{dsfont}
\usepackage{wrapfig}
\usepackage{xspace}
\usepackage[T1]{fontenc}
\usepackage{times}
\usepackage{cite}
\usepackage{tikz}
\usepackage{subcaption}
\usetikzlibrary{}
\usetikzlibrary{shapes,arrows, chains, decorations.pathmorphing}
\usepackage[font=footnotesize]{caption}

\usetikzlibrary{decorations.markings}
\tikzset{
    set arrow inside/.code={\pgfqkeys{/tikz/arrow inside}{#1}},
    set arrow inside={end/.initial=>, opt/.initial=},
    /pgf/decoration/Mark/.style={
        mark/.expanded=at position #1 with
        {
            \noexpand\arrow[\pgfkeysvalueof{/tikz/arrow inside/opt}]{\pgfkeysvalueof{/tikz/arrow inside/end}}
        }
    },
    arrow inside/.style 2 args={
        set arrow inside={#1},
        postaction={
            decorate,decoration={
                markings,Mark/.list={#2}
            }
        }
    },
}

\tikzset{snake it/.style={decorate, decoration=snake}}

\usepackage[noend]{algpseudocode}
\usepackage{algorithm}
\algnewcommand{\LeftCComment}[1]{\fontsize{5}{10}\Statex \(//\) #1}
\algnewcommand{\CComment}[1]{\fontsize{5}{10}\hskip3em\(//\)\fontsize{5}{10} #1}
\algrenewcommand\algorithmicindent{1.0em}%

\newcommand\blfootnote[1]{%
  \begingroup
  \renewcommand\thefootnote{}\footnote{#1}%
  \addtocounter{footnote}{-1}%
  \endgroup
}

\begin{document}

\title{Collaborative Dense SLAM}

\author[]{Louis Gallagher
%\thanks{This research was supported, in part, by the IRC GOIPG scholarship scheme grant GOIPG/2016/1320 and, in part, by Science Foundation Ireland grant 13/RC/2094}
}
\author[]{John B. McDonald}
%\author{Anonymous}
\affil{Department of Computer Science, Maynooth University, Maynooth, Co. Kildare, Ireland}
\affil{louis.gallagher.2013@mumail.ie, johnmcd@cs.nuim.ie}
%\affil{}
%\affil{Anonymous}

\date{}
\maketitle
\thispagestyle{empty}

\begin{abstract}
%Collaborative visual SLAM is an extension of the standard visual SLAM problem whereby the goal is to perform the reconstruction using multiple independently moving cameras. . While collaborative mapping has been a recurring theme in the SLAM literature over the years, in terms of the move towards dense reconstructions it has not recieved the same focus as single-sensor visual SLAM.\par 
%\blfootnote{This research was supported, in part, by the IRC GOIPG scholarship scheme grant GOIPG/2016/1320 and, in part, by Science Foundation Ireland grant 13/RC/2094}
\blfootnote{This research was supported, in part, by the IRC GOIPG scholarship scheme grant GOIPG/2016/1320 and, in part, by Science Foundation Ireland grant 13/RC/2094 to Lero - the Irish Software Research Centre (www.lero.ie)}

In this paper, we present a new system for live collaborative dense surface reconstruction. Cooperative robotics, multi participant augmented reality and human-robot interaction are all examples of situations where collaborative mapping can be leveraged for greater agent autonomy. Our system builds on ElasticFusion to allow a number of cameras starting with unknown initial relative positions to maintain local maps utilising the original algorithm. Carrying out visual place recognition across these local maps the system can identify when two maps overlap in space, providing an inter-map constraint from which the system can derive the relative poses of the two maps. Using these resulting pose constraints, our system performs map merging, allowing multiple cameras to fuse their measurements into a single \emph{shared} reconstruction. The advantage of this approach is that it avoids replication of structures subsequent to loop closures, where multiple cameras traverse the same regions of the environment. Furthermore, it allows cameras to directly exploit and update regions of the environment previously mapped by other cameras within the system.\par
%  provides us with a mechanism for aligning and fusing the maps into a common global reference frame. Cameras that have been aligned in this way can proceed to fuse measurements into, and track against, the single shared map.\par

%Inevitably the system will fail to align certain of the submaps, this may be due to either visual place recognition failing to find visual overlap or to the system failing to derive the relative transformation. In such cases we use global registration techniques to derive a single global coordinate frame directly.\par
 
We provide both quantitative and qualitative analyses using the syntethic ICL-NUIM dataset and the real-world Freiburg dataset including the impact of multi-camera mapping on surface reconstruction accuracy, camera pose estimation accuracy and overall processing time. We also include qualitative results in the form of sample reconstructions of room sized environments with up to $3$ cameras undergoing intersecting and loopy trajectories. \par

\end{abstract}
\textbf{Keywords:} $3D$ reconstruction, Dense Mapping, Collaborative Mapping, SLAM, Machine Vision

%%%%%%%%%%%%%%%%%%%%%%
\section{Introduction}
%\begin{wrapfigure}{r}{0.5\textwidth}
  %\vspace{-20pt}
  %\begin{center}
   % \includegraphics[width=0.4\textwidth, height=0.28\textwidth]{belfast_campus.jpg}
%\end{center}
%\vspace{-20pt}
  %\caption{Introduction Banner.}
  %\vspace{-0pt}
%\end{wrapfigure}
%%%%%%%%%%%%%%%%%%%%%%
\textit{Simultaneous localisation and mapping} (SLAM) is a core problem in robotics and computer vision, 
%and a problem that precedes
providing the foundations for many higher level environment understanding and interaction tasks, particularly in autonomous mobile robotics and augmented reality. With the advent of consumer grade active depth sensors, such as the Microsoft Kinect and the Asus Xtion Pro, and the additional compute power made available through modern GPUs, dense mapping has become a prominent focus of research in visual SLAM. Recently, the state-of-the-art in the field has come to include a multitude of systems offering large-scale, high-precision mapping and tracking capabilities \cite{bundlefusion, elasticfusionjournal, dvo, kintinuous, kinectfusion, lsdslam, orbslam2, dtam, pbfusion}. What distinguishes dense mapping from its sparse counterpart is twofold; firstly, a direct approach to camera tracking where potentially every pixel is used; secondly, the estimation of a map that consists of a dense set of measurements of the surface that underlies the scene.
%, as opposed to the estimation of a sparse set of geometric primitives that lie on that surface.
 Examples of such map representations include volumetric signed distance functions and surface element (surfel) lists.\par

Much of the developments within dense SLAM have been focused on scenarios involving mapping the environment using a single-sensor platform. However, many scenarios exist where multiple mobile agents must work together in the same space,
% each one requiring a map of the world around it.
thereby requiring a single shared map of the environment. The main contribution of this paper is a system that addresses the wider problem of online collaborative, multi-camera dense mapping and tracking using RGB-D sensors with unknown initial relative poses.\par 

Historically collaborative mapping has been a recurring theme in the SLAM literature where many problems that emerge when multiple cameras must map collaboratively have been addressed \cite{multislamreview, cooperative, submapcollab, sparseeif, particlefilter, manifold, rendezvous, multisession}. For example, in general the starting position of cameras relative to one another is not known and so they do not share a common reference frame, which instead must be identified online, as mapping proceeds. Once a common reference frame has been found maps must be merged, from which point updates to the map from multiple cameras must be coordinated so as to avoid corrupting the map. \par
%It is also necessary to consider the manner in which a collaborative mapping systems performance can be quantified and compared to the performance of other collaborative and single-camera mapping systems. Although many datasets are available for benchmarking single camera mapping few exist that can be used to measure the performance of a collaborative system.\par

%One must consider additional complexities relating to what data is distributed across the team of agents and whether or not that data should be processed in a decentralised or centralised manner.
%These complexities themselves entail further questions around how data is communicated and the robustness and availability of the communication channels being used.
% In our system all mapping and tracking operations are carried out at a central server and we assume that frames from different cameras are sent to this server over an idealised communication channel. Although we do acknowledge the complexities of real-world data communication, making this simplifying assumption allows us to focus on the specific question of collaborative mapping.\par

Collaborative mapping has come back into focus in SLAM research in recent times. Golodetz et al. propose a novel approach to dense collaborative mapping that uses InfiniTAM \cite{InfiniTAM_V3_Report_2017} to construct \emph{separate} voxel-based maps for each camera. The system uses a random forest relocaliser to find estimates of the relative transforms between each of these camera specific maps. These relative transforms act as constraints in a pose graph which, once optimised, yields a set of optimal global poses, one for each of the sub maps. %The system can run in `interactive' mode, where it attempts to align the maps online, or batch mode, where it attempts to align the maps once the cameras have stopped exploring.
Although these global poses align the submaps into a common reference frame they are never merged directly, instead each camera maintains its own submap throughout \cite{collabdense}.
%In this way the system can distribute the mapping of large spaces across a team of cameras \cite{collabdense}.
 In contrast, our system performs map merging, allowing multiple cameras to fuse their measurements into a single \emph{shared} reconstruction. The advantage of this approach is that it avoids replication of structures subsequent to loop closures, where multiple cameras traverse the same regions of the environment. Furthermore, it allows cameras to directly exploit and update regions of the environment previously mapped by other cameras within the system.\par % is aimed at allowing multiple cameras to reconstruct a single \emph{shared} model.\par 

The remainder of this paper is structured as follows. In Section \ref{ef} we briefly review the ElasticFusion (EF) dense mapping system. In Section \ref{dcm} we describe how we extended EF to allow multi-camera mapping. In section \ref{intermap} we explain the inter-map loop closure mechanism and how maps are merged. Later, in Section \ref{eval} we describe the experimental process used to measure the systems performance and present the results of this experimentation. Finally, in Section \ref{conclusion} we conclude with some remarks on the proposed system and give some directions for future research.\par
%%
%The attributes of position, normal, radius and colour describe a surfel's physical location, orientation, scale and colour.
\section{Background}\label{ef} 
In this section we briefly summarise the main elements of the ElasticFusion (EF) algorithm and define the notation used for the remainder of the paper. For a more comprehensive treatment of EF see %\cite{elasticfusion} and 
\cite{elasticfusionjournal}. 
%EF conforms to the \textit{dense alternation} paradigm of mapping. In this paradigm processing of each frame is broken into two distinct phases; a tracking phase and a mapping phase. In the tracking phase the map is held fixed and used to estimate the pose of the camera for the current frame. Then, in the mapping phase, the newly estimated pose of the camera is used to fuse the current frame into the map. 
In EF the map is represented as an unordered list of surface elements (surfels), $\mathcal{M}$. Each surfel is an estimate of a discrete patch centred around an associated scene point. A surfel is characterised by its position $p \in \mathbb{R}^3$, normal $n \in \mathbb{R}^3$,  radius $r \in \mathbb{R}$, weighting $w \in \mathbb{R}$, colour $c \in \mathbb{N}^3$, initialisation time $t_0$ and the time it was last observed $t$. The weight of a surfel signifies the level of confidence in the estimation of its parameters. $\mathcal{M}$ is split along temporal lines into two non-intersecting sub-lists, $\Theta$ containing surfels that are considered active, meaning they have recently been observed by the camera, and $\Phi$ containing those surfels that have not been observed recently and are considered inactive. %EFs main processing pipeline consists of four main elements which we will now describe.\par % A depiction of the system from an information flow point of view can be seen in Figure \ref{pipeline}.\par 

%Frames are processed sequentially, in the order that they are captured and one at a time. Initially, the camera's centre of projection is assumed to be at the origin of the global coordinate frame with it's \textit{principal axis} (i.e. $Z$-axis) coincident with the $Z$-axis of the global coordinate frame. 

Camera tracking is achieved by aligning each input frame at time $t$ with a predicted view of the model which is rendered from $\Theta$ using the pose $P_{t-1}$.
%Camera tracking is the first stage of the pipeline, the aim of which is to compute an initial estimate of the 6 \textit{degree of freedom} (DOF) pose, $\hat{P}_t \in \mathbb{SE}_3$, of the camera for the current frame. To track the camera a predicted view of the model is rendered from $\Theta$ using $P_{t-1}$, the pose of the camera at the previous timestep, to define the image plane. The predicted frame is aligned to the current live frame by finding an estimate, $\hat{T} \in \mathbb{SE}_3$, of the rigid body transformation between them. 
The transformation that brings the two frames into alignment is estimated by minimising a combined photometric and geometric error over a three level Gaussian pyramid.% to ensure that in cases of large inter-frame displacement the minimisation will still converge. 
The resulting transformation is applied to $P_{t-1}$ to yield an estimate of the global pose of the camera, $P_{t}$, for the current time step. \par

\tikzstyle{decision} = [diamond, draw, fill=blue!20, 
    text width=4.5em, text badly centered, node distance=3cm, inner sep=0pt]
\tikzstyle{block} = [rectangle, draw, fill=blue!20, 
    text width=5em, text centered, rounded corners, minimum height=4em]
\tikzstyle{line} = [draw, -latex']
\tikzstyle{cloud} = [draw, ellipse,fill=red!20, node distance=3cm,
    minimum height=2em]
\tikzstyle{coord} =[draw, coordinate, node distance=6mm and 25mm]

Following camera tracking, in the next stage of the pipeline EF seeks to close two types of loop; global and local loops. Global loop closures are identified with a fern-based visual place recognition system as proposed by \cite{fern}. %In brief a fern encoding of an RGB-D image is a binary code which allows for frames to be efficiently compared with one another, where the similarity of frames is measured in terms of the hamming distance between their respective codes. Incoming frames that are sufficiently dissimilar to all previously encoded frames (i.e. the hamming distance is large) are harvested and kept in the database as keyframes. Similarly, where the hamming distance between an incoming frame and a frame in the fern database is low a potential loop closure has been found.\par% To encode frames a \textit{conservatory} of ferns are assigned  fixed locations uniformly sampled from the image domain. A fern consists of $4$ nodes, one for each image channel, and each node in turn carries a randomly sampled binary threshold parameter. The conservatory of ferns is embedded into incoming frames and the thresholded response at each node of each fern is computed resulting in a series of binary codes, one for each fern. These binary codes are used to index into a look-up table that maps from fern code blocks to the ids of frames that have the same fern code. This table along with the corresponding frame poses constitutes what is referred to as a fern-encoding database. One can think of this process as being akin to feature extraction and matching, where the fern codes correspond to features and the table look-up corresponds to feature matching. By counting the number of co-occurring blocks one can compute the block-wise hamming distance between frames, or between a frame and a set of frames, and can thus get a measure of the similarity between them.  
Using the pose $P_{t}$, from the camera tracking step, a new active model prediction is rendered from which a fern encoding is computed and used to search the fern database for a matching view. If one is found then the two views are aligned,
%, bringing the active model prediction into alignment with the matched fern. The final transformation from this registration is used to 
generating a set of surface-to-surface constraints which are used to optimise the nodes in a deformation graph. 
The nodes of this graph are initialised by subsampling a set of surfels from $\mathcal{M}$. Once optimised, the graph can be applied directly to the surface. 
To do this an influence region is computed for each surfel consisting of the set of graph nodes close both spatially and in initialisation time.
%nodes nearest to it in terms of initialisation time are found, from which the closest nodes to it in space are taken. This set forms the influence region for that surfel. 
 The final pose of a surfel is given by the weighted application of the affine transformations held in each of the nodes in the surfel's influence region where the weight of a given node is a function of its distance from the surfel.
 %, with nodes that are further away having less influence.\par %A deformation is accepted if two conditions are met: (i) the alignment between the current model active prediction and the matched fern is successful; (ii) the deformation requires a significant realignment of the maps geometry and that realignment is consistent with the maps geometry as defined by the cost function for the deformation grap optimisation. If the deformation is accepted then the pose of the camera is updated as per the registration with the fern view. \par
 
%\subsection{Local Loop Closures}
If a global loop closure has not occurred at the current time step then local loop closures aligning $\Theta$ to $\Phi$ are sought.
%The result of a local loop closure is the alignment of $\Theta$ and $\Phi$ and the reactivation of inactive surfels.
 Local loops are closed by performing a surface registration between the portions of $\Theta$ and $\Phi$ in view of $P_t$. Similar to global loop closures this registration yields a set of surface-to-surface constraints that are used to optimise the nodes of a deformation graph. Once the deformation graph has been applied the surfels in $\Phi$ that were in view are reactivated within $\Theta$.\par %and active portions of the map are brought into alignment.\par % and as before the pose of the camera iupdated in accordance with its registration with the inactive view.\par 

%\subsection{Fusion} 
Fusion of the current frame can proceed once the camera has been tracked and all loop closures have been resolved. Surfels in $\Theta$ are projectively associated with pixels in the current frame using $P_t$ and $K$, the camera's intrinsic parameters. The normals, vertices, colours and radius of associated points are merged with a weighted average of the model surfel and the newly measured vertex. The weight of the surfel is updated in accordance with the normalised radial distance of the measured vertex from the camera centre. If there is no corresponding surfel for a vertex then a new unstable surfel is inserted into the model. Surfels become stable after their confidence counter reaches a threshold through repeated measurement, from which point they can be used for camera tracking.\par

\section{Approach}\label{dcm}

In this section we 
%propose an approach to 
extend the EF system to permit dense mapping using multiple independently moving RGB-D cameras to map a space in a collaborative fashion. As the cameras explore a space they fuse their measurements into, and track against, a shared model of the reconstruction.% Taking the ElasticFusion dense mapping system that we described in Section \ref{ef} as our starting point we extend the system such that it can accommodate multi-camera mapping and tracking. \par%The main idea behind our approach is for each camera to initially maintain its own local map. We then seek out inter-map loop closures between these camera specific local maps\par

%\addcontentsline{toc}{chapter}{Dense Collaborative Mapping}
%\section{Extending ElasticFusion for Multi-Camera SLAM}\label{approach}
%\begin{figure}[!htb]
%\begin{minipage}{1.0\linewidth}
%\centering
%\subfloat[][]{\includegraphics[width=0.45\linewidth]{images/ef_co_01}\label{ef1}}
%\hfill
%\subfloat[][]{\includegraphics[width=0.45\linewidth]{images/ef_co_02}\label{ef2}}
%\end{minipage}

%\begin{minipage}{1.0\linewidth}
%\centering
%\subfloat[][]{\includegraphics[width=1.0\linewidth, height = 6cm, keepaspectratio]{images/ef_co_03}\label{ef3}}
%\end{minipage}
%\caption[A two camera collaborative mapping session in our system.]{A two camera collaborative mapping session in our system. All cameras started in general position with no \textit{a priori} knowledge of the relative poses between them. \protect\subref{ef1} camera one and \protect\subref{ef2} camera two compute local maps using EF. \protect\subref{ef3} A fern based visual intermap loop closure allows the local maps of \protect\subref{ef1} and \protect\subref{ef2} to be merged.}
%\label{fig:comap}
%\end{figure}

In extending EF to permit collaborative mapping we identify two distinct phases of processing for any given input stream. 
%This distinction arises from the fact that in collaborative mapping there is, in general, no common frame of reference to begin with, and hence each camera's initial pose is unknown relative to the other cameras. 
In the initial phase, the system assumes that each camera is positioned at the origin of a frame of reference that is independent to that of other cameras essentially processing it via an independent EF mapping pipeline. As mapping progresses the aim is for inter-camera loop closures to occur, thereby providing the necessary transformations between the camera submaps. These transformations permit alignment of submaps into a common frame of reference which makes it possible to fuse subsequent measurements from each camera into a single global map.\par 

A set of $n$ cameras $\{\mathds{C}_i\}, i \in [1..n]$ serve frames to a central server that runs our mapping process. Each $\mathds{C}_i$ is represented by a $4 \times 4$ transformation matrix from the Lie group $\mathds{SE}_3$. %For the purposes of this work we assume frames are sent over a network of sufficient bandwidth such that the network does not form a bottleneck in the system. However, in practise 
 We use the Lightweight Communications and Marshalling (LCM) system of \cite{lcm} to connect the cameras to the central mapping server over a network. LCM provides efficient encoding and decoding mechanisms for transmission of camera data over a network and the susbsequent unpacking of that data into data structures that are more amenable to processing.\par 

Given that there are multiple cameras each frame is of the form $\mathds{F}_i^t$ where the superscript $t$ indicates the frame's timestamp and the subscript $i$ denotes that this frame is from camera $\mathds{C}_i$. We note that the cameras need not be synchronised in time.\par% Indeed, the system can be used in a multi-session capacity where multiple explorations of the same environment are processed back to back.\par % Furthermore, all of the cameras are assumed to be synchronised in time, meaning frame $\mathds{F}_i^t$ was captured from camera $\mathds{C}_i$ at the same time that frame $\mathds{F}_j^t$ was captured from camera $\mathds{C}_j$.\par

If we assume momentarily that all cameras are globally aligned then multi-camera fusion proceeds in essentially the same way as in the original EF algorithm. At each timestep $t$ the frames from each camera for that timestep are processed in \emph{sequence} by EF. That is, each frame is processed in full by EF before the next camera's frame for that timestep is processed. Note that the ordering of the cameras here is arbitrary.\par 

An important difference between our system and EF is that each camera has its own active and inactive map regions, $\{\Theta _i\}$ and $\{\Phi _i\}$. %there are now multiple, potentially intersecting, active map regions, $\{\Theta _i\}$, one for each camera. Thus there are multiple inactive regions, $\{\Phi _i\}$, again, one for each camera.
 This allows cameras to close local loops in situations where they transition to regions of the map that are in the active regions of other cameras but that are inactive with respect to itself. To achieve this the previously scalar last seen time attribute of a surfel takes the new form of an array $t = \{t_0, ..., t_n\}$ where $t_i$ contains the time that the surfel was last seen by camera $\mathds{C}_i$, for $ i \in [1..n]$. When a measurement from frame $\mathds{F}_i^t$ is fused into surfel $\mathcal{M}^s$ then the last seen time of that surfel is updated as $\mathcal{M}^s_{t_i} = t$. Each camera uses its own active and inactive regions for mapping and tracking. For global loop closure detection, all cameras in a given \textit{reference frame} (see below) collectively maintain a single Fern database which they use to independently check for and close global loops.\par

%\begin{equation}
%	\mathcal{M}^s_{t_i} = t
%\end{equation}	

%\vspace{-22pt}

%\par

Returning now to the initial phase of processing where cameras are operating independently and the transformations of each camera relative to a shared coordinate frame are unknown. Periodically, loop closures between camera specific submaps are sought. In Section \ref{intermap} we outline our strategy for determining inter-map loop closures. Once a loop between two submaps has been identified and the transformation between them computed the two submaps can be merged into a single map from which point multi-camera fusion can proceed as described above. We capture this situation with a \textit{reference frame}, which is a tuple of the form $\mathcal{R}_j = \{\mathcal{M}_j, \{\mathds{C}_i\}\}$, for surfel map $\mathcal{M}_j$ and set of cameras $\{\mathds{C}_i\}$ which incrementally update and track against $\mathcal{M}_j$ through multi-camera fusion. Additionally each $\mathcal{R}_j$ maintains its own fern database for inter and intra-map loop closures.\par

\subsection{Fern-based Inter-Map Loop Closures}\label{intermap}

Our approach to inter-map loop closures is an extension of EF's existing intra-map global loop closure mechanism described in Section \ref{ef}. Each camera is initialised in its own reference frame. For each current active frame prediction $\mathds{F}_i^t$ we search the fern databases of all the other reference frames for a matching view. If one is found then we compute the transformation between the reference frame of the corresponding camera, $\mathds{C}_i$, and the reference frame of the matched fern using the camera odometry from Section \ref{ef}. For example, if we find a loop closure between reference frames $\mathcal{R}_j$ and $\mathcal{R}_k$ by matching frame $\mathds{F}_i^t$, captured by $\mathds{C}_i^j$ in $\mathcal{R}_j$, to a fern in the fern encoding database of $\mathcal{R}_k$ then the result of aligning the fern and the frame is a transformation $H_i^f \in \mathds{SE}_3$ that aligns the matched fern and $\mathds{F}_i^t$. For clarity, we have added the superscript $j$ to the camera, which denotes the frame of reference of the camera. $H_i^f$ gives an initial estimate of the transformation that aligns $\mathcal{R}_j$ with $\mathcal{R}_k$ as follows

\begin{equation}
 \hat{T}_j^k = H_i^f(\mathds{C}_i^j)^{-1}
\end{equation}

To further refine the transformation and verify that it is valid we first predict a full resolution frame, $_k\mathds{F}_i^t$ using $\mathcal{M}_k$ and $\mathds{C}_i^k$, which is the transformed pose of camera $\mathds{C}_i$ in frame of reference $\mathcal{R}_k$.
%\begin{equation}
% \mathds{C}_i^k = \hat{T}_j^k\mathds{C}_i^j
%\end{equation}
 That is, we predict the view that $\mathds{C}_i$ has of $\mathcal{M}_k$. Taking this predicted frame we align it to the original frame $\mathds{F}_k^t$, again using the camera odometry described in Section \ref{ef}. The result of this is a refined estimate, $T_j^k$, of the transformation from $\mathcal{R}_j$ to $\mathcal{R}_k$. Before accepting the loop closure as a valid one we inspect the final error and inlier count of the camera tracking cost function. For the loop closure to be accepted the error must be low while the number of inliers must be high. Inliers here refer to depth measurements in $\mathds{F}_i^t$ which have been associated with a corresponding depth value in $_k\mathds{F}_i^t$. As per the original EF algorithm the final state of the cost function covariance matrix is also checked to ensure that the cost function was sufficiently constrained along each of the $6$ degrees of freedom of the camera. If all these conditions are met then the inter-map loop closure is accepted and the maps are merged.\par

To merge the two maps we apply $T_j^k$ to the positions and normals of all of the surfels $ \mathcal{M}^s \in \mathcal{M}_j$ and all of the camera poses, $\{\mathds{C}_i\}$ in frame of reference $\mathcal{R}_j$ as follows 

\noindent\makebox[\textwidth][c]{%
\noindent\begin{minipage}{.33\linewidth}
\begin{equation}
\mathcal{M}^s_{k_p} = T_j^k\mathcal{M}^s_{j_p}\hspace{-3.5em}
\end{equation}
\end{minipage}
\begin{minipage}{.33\linewidth}
	\begin{equation}
    \mathcal{M}^s_{k_n} = R_j^k\mathcal{M}^s_{j_n}\hspace{-3.5em}
	\end{equation}
\end{minipage}
\begin{minipage}{.33\linewidth}
\begin{equation}
\mathds{C}_i^k = T_j^k\mathds{C}_i^j \hspace{-3.5em}
\end{equation}
\end{minipage}}

%\noindent\makebox[\textwidth][c]{%
%\noindent\begin{minipage}{.33\linewidth}
%\begin{equation}
%\mathds{C}_i^k = T_j^k\mathds{C}_i^j \hspace{-8.5em}
%\end{equation}
%\end{minipage}}
%\begin{flalign}
%	\mathcal{M}^s_p &= T_j^k(\mathcal{M}^s_p,1)\\
%	\mathcal{M}^s_p &= R_j^k(\mathcal{M}^s_n,1)\\
%	\mathds{C}_i^k &= T_j^k\mathds{C}_i^j 
%\end{flalign}

where $R_j^k \in \mathds{SO}_3$ is formed from the upper left $3 \times 3$ submatrix of $T_j^k$. Note that we have not explicitly shown conversions to and from homogenous representations for the sake of brevity. This process is visualised in Figure \ref{fern}. The algorithm is listed in Algorithm \ref{alg:dcm}.

\begin{figure}
%\vspace{-2.5cm}
%\centering
\begin{minipage}{0.45\linewidth}
%\vspace{-5cm}
\begin{algorithm}[H]  
  \scriptsize
	\caption[Dense collaborative mapping.]{Dense collaborative mapping.}
	\label{alg:dcm}
	\begin{algorithmic}[1]
		\Procedure{CollaborativeMapping}{$c$}
			\Statex \textbf{Input:} a set of $n$ cameras $\{\mathds{C}_i\}$
			\Statex \textbf{Output:} a set of $m$ maps $\{\mathcal{M}_j\}$ where $m \leq n$
			\Statex
			
			%\LeftCComment{initialise a new map for each camera.}
			\State Dict $c\_to\_m \gets \emptyset$
			
			\For{ $\mathds{C}_i \in c$}
				\State $c\_to\_m[\mathds{C}_i] \gets$ new Map() 
			\EndFor
			
			\Statex
			
			\Loop
				%\LeftCComment{process the next frame from each camera.}
				\For{$\langle\mathds{C}_i,\mathcal{M}_j\rangle$ $\in$ $c\_to\_m$}
					\State map\_and\_track($\mathds{C}_i$, $\mathcal{M}_j$)
				\EndFor
				
				\Statex
				
				%\LeftCComment{Intermap loop closures.}
				\For{$\mathds{C}_i \in c\_to\_m$}
					\State Map $\mathcal{M}_k \gets c\_to\_m[\mathds{C}_i]$
					\State Mat$4 \times 4$ $P_i^k \gets \mathds{C}_i.currentPose()$ 
					\State Frame $\mathds{F}_{\mathds{C}_i^k}^{p} \gets$ predictView($P_i$, $\mathcal{M}_j$) 
					\For{$\mathcal{M}_j \in c\_to\_m$}
						\If{$\mathds{C}_i \notin \mathcal{M}_j$}
							\State Mat4$\times$4 $P_i^k \gets$ findFern($\mathds{C}_i$, $\mathcal{M}_j$)
							 %\CComment{\parbox[t]{.3\linewidth}{$\mathds{C}_i$ in $\mathcal{M}_j$'s coordinate frame.}}
							\State Frame $\mathds{F}_{\mathds{C}_i^j}^{p} \gets$ predictView($P_i^j$, $\mathcal{M}_j$)
							 %\CComment{\parbox[t]{.3\linewidth}{predict view of $\mathds{C}_i$ in $\mathcal{M}_j$}}
							\State $P_i^j \gets $rgbdOdometry($\mathds{F}_{\mathds{C}_i^j}^{p}$, $\mathds{F}_{\mathds{C}_i^k}^{p}$, $P_i^j$)
							%\CComment{\parbox[t]{.3\linewidth}{align predicted views.}}
							\Statex
							\If{rgbdOdometry was successful}
								\State Mat4$\times$4 $T_k^j \gets PP^{-1}_i$
								\State $\mathcal{M}_j \gets$merge($\mathcal{M}_j$, $\mathcal{M}_k$, $T_k^j$)
								\State $c\_to\_m[\mathds{C}_i] \gets \mathcal{M}_j$
							\EndIf						
						\EndIf
					\EndFor
				\EndFor
			\EndLoop
		\EndProcedure
	\end{algorithmic}
\end{algorithm}
\end{minipage}%
\hfill
\begin{minipage}{0.45\linewidth}
\begin{tikzpicture}[x=1cm,y=1cm,z=1cm,>=stealth, scale=0.6, every node/.style={scale=0.6}]
  \draw [blue] decorate [decoration={zigzag}] {(-4, -4) -- (6.6,-4)};
  \coordinate[label= above left:$\mathcal{M}_1$] (m) at (6.6,-4);
  \draw[->] (xyz cs:x=-4.4,y=-4.4) -- (xyz cs:x=-4,y=-4.4) node[below] {\tiny $X$};
  \draw[->] (xyz cs:x=-4.4,y=-4.4) -- (xyz cs:x=-4.4,y=-4) node[above] {\tiny $Y$};
  
  \node[fill=black,regular polygon, regular polygon sides=3,inner sep=2.5pt] at (3,-3.5) {};
  \node[fill,circle,inner sep=1.5pt] at (3,-3.3) {};
	\coordinate[label= above left: \text{\tiny $\mathds{C}_1$}] (c1) at (3,-3.5);
	
	\draw [black] plot [smooth, tension=1] coordinates { (-4.4,-4.4) (-0.7, -2.95) (3,-3.5)} [arrow inside={end=stealth,opt={red,scale=2}}{0.55}];  
  
  \draw [green] decorate [decoration={zigzag}] {(-4, -1) -- (6.6,-1)};
  \coordinate[label= above left:$\mathcal{M}_2$] (m_2) at (6.6,-1);
   \draw[->] (xyz cs:x=-4.4,y=-1.4) -- (xyz cs:x=-4,y=-1.4) node[below] {\tiny $X$};
  \draw[->] (xyz cs:x=-4.4,y=-1.4) -- (xyz cs:x=-4.4,y=-1) node[above] {\tiny $Y$};
  
   \node[fill=black,regular polygon, regular polygon sides=3,inner sep=2.5pt] at (2,-0.5) {};
   \node[fill,circle,inner sep=1.5pt] at (2,-0.3) {};
   
   \node[draw=purple, dashed, regular polygon, regular polygon sides=3,inner sep=2.5pt, rotate around={-15:(0,0)}] at (4.5,-0.1) {};
   \node[fill,circle,inner sep=1.5pt] at (4.55, 0.11) {};
	\coordinate[label= above left: \text{\tiny $\mathds{C}_f$}] (c1) at (2,-0.5);
	\coordinate[label= above right: \text{\tiny $\mathds{C}_1^\prime$}] (c1) at (4.5,-0.1);
		\draw [black] plot [smooth, tension=1] coordinates { (-4.4,-1.4) (-0.7, -0.05) (2,-0.5)} [arrow inside={end=stealth,opt={red,scale=2}}{0.7}];
		\draw [black] plot [smooth, tension=1] coordinates { (2,-0.5) (3.5, 0.05) (4.5,-0.1)}[arrow inside={end=stealth,opt={red,scale=2}}{0.61}];
		\draw [black] plot [smooth, tension=1] coordinates { (-4.4,-1.4) (0.0, 1.1) (4.5, -0.1)}[arrow inside={end=stealth,opt={red,scale=2}}{0.55}];

 \draw [black] plot [smooth, tension=1] coordinates { (3.0,-3.5) (4.75, -1.75) (4.5, -0.1)} [arrow inside={end=stealth,opt={red,scale=2}}{0.6}];
		
 \coordinate[label= below left: \text{\tiny $T_2^{\mathds{C}_f}$}] (t2f) at (0.5,0.02);
 \coordinate[label= above left: \text{\tiny $T_2^{\mathds{C}_1^\prime}$}] (t2p) at (0.5,1.1);
 \coordinate[label= below left: \text{\tiny $T_{\mathds{C}_f}^{\mathds{C}_1^\prime}$}] (tfp) at (3.7,0.05);
 \coordinate[label= above left: \text{\tiny $T_1^{\mathds{C}_1}$}] (t1c) at (-0.7,-2.95);
  \coordinate[label= right: \text{\tiny $T_{\mathds{C}_1}^{\mathds{C}_1^\prime}$}] (t1c) at (4.75,-1.75);
  
  \draw [red, dashed] (-5.0, -5.4) -- (7,-5.4);
  \coordinate[label = above:\text{pre loop closure}] (d) at(5.2, -5.2);
  \coordinate[label = below:\text{post loop closure}] (d) at(5.2, -5.6);
  
  \draw [green] decorate [decoration={zigzag}] {(-4, -8.0) -- (1.3,-8.0)};
  \draw [blue] decorate [decoration={zigzag}] {(1.3,-8.0) -- (6.6,-8.0)};
  \coordinate[label= above left:$\mathcal{M}_1 \cup \mathcal{M}_2$] (m) at (6.6,-7.6);
  \draw[->] (xyz cs:x=-4.4,y=-8.4) -- (xyz cs:x=-4,y=-8.4) node[below] {\tiny $X$};
  \draw[->] (xyz cs:x=-4.4,y=-8.4) -- (xyz cs:x=-4.4,y=-8.0) node[above] {\tiny $Y$};
  \node[fill,circle,inner sep=1.5pt] at (4.57,-7.25) {};
  \node[draw=black, fill, regular polygon, regular polygon sides=3,inner sep=2.5pt, rotate around={-15:(0,0)}] at (4.5,-7.5) {};
  \draw[black] plot [smooth, tension = 1] coordinates{(-4.4,-8.4) (0, -6.5) (4.5, -7.5)} [arrow inside = {end=stealth, opt={red,scale=2}}{0.6}];
  \coordinate[label= above left: \text{\tiny $T_2^{\mathds{C}_1}$}] (t2c1) at (1,-6.5);
  \coordinate[label= above left: \text{\tiny $\mathds{C}_1^2$}] (t2c12) at (4.57,-7.35);
\end{tikzpicture}
\caption[Fern based intermap loop closure.]{A fern based intermap loop closure. Camera $\mathds{C}_1$ matches its current view of $\mathcal{M}_1$ to a fern view, denoted by $\mathds{C}_f$, in $\mathcal{M}_2$'s fern database. The surfaces in each view are geometrically consistent and so a loop is closed between $\mathcal{M}_1$ and $\mathcal{M}_2$. The loop closure itself results in $T_2^{\mathds{C}_1^\prime} = T_2^{\mathds{C}_f}T_{\mathds{C}_f}^{\mathds{C}_1^\prime}$, the pose of $\mathds{C}_1$ in the reference frame of map $\mathcal{M}_2$.  Applying the composite transformation, $T_{\mathds{C}_1}^{\mathds{C}_1^\prime} = T_2^{\mathds{C}_1^\prime}(T_1^{\mathds{C}_1})^{-1}$, brings the pose $\mathds{C}_1$ and the surfels $\mathcal{M}_1$ into the coordinate frame of $\mathcal{M}_2$, resulting in the fused map.}
%Composing transformations as follows yields a transformation, $T_{\mathds{C}_1}^{\mathds{C}_1^\prime}$, that brings the pose of $\mathds{C}_1$ and the surfels of $\mathcal{M}_1$ into the coordinate frame of $\mathcal{M}_2$; 
%let $T_2^{\mathds{C}_1^\prime} = T_2^{\mathds{C}_f}T_{\mathds{C}_f}^{\mathds{C}_1^\prime}$ be the pose of $\mathds{C}_1$ in  as before, then the transformation 
%$T_{\mathds{C}_1}^{\mathds{C}_1^\prime} = T_2^{\mathds{C}_1^\prime}(T_1^{\mathds{C}_1})^{-1}$ }
\label{fern}
\end{minipage}%
%\caption{osijv}
\end{figure}

\section{Experimental Evaluation}\label{eval}

In benchmarking our system we aim to do two things: (i) quantitatively measure the impact of multiple sensors on mapping and tracking performance in terms of both accuracy and processing time; (ii) qualitatively demonstrate the capability of our system. To achieve this we use two standard SLAM benchmark datasets, the ICL-NUIM dataset of \cite{iclnuim} and the RGB-D dataset of \cite{freiburg}. All experiments were run on a machine equipped with an NVidia GeForce GTX 1080 Ti GPU, 11GB of GDDR5 VRAM, a 4 core Intel i7-7700K CPU running at 4.20GHz and 16GB of DDR4 system memory. For a visualisation of dense collaborative mapping please see the accompanying video (\url{https://youtu.be/GUtHrKEM85M}).\par

We performed a quantitative analysis of the performance of the system using the living room scene from the ICL-NUIM dataset.
% of \cite{iclnuim}. %This dataset contains two scenes, the living room scene and the office scene.
 There are four trajectories through this scene accompanied by both ground truth camera poses for each trajectory and a synthetic ground truth surface model. In our experiments, we used three of the four living room trajectories, $kt_0$, $kt_1$ and $kt_2$. We note that we have not included $kt_3$ in our results due to the fact that $kt_3$ exposes a failure mode of the system. $kt_3$ is one of the more challenging sequences in the ICL-NUIM dataset, as is reflected in the results of both \cite{bundlefusion} and \cite{elasticfusionjournal}. We found that including $kt_3$  in any of the combinations in our experiments resulted in significantly reduced reconstruction quality.   
% We left the fourth sequence, $kt_3$, out since including this sequence led to poor surface reconstructions.
We are currently investigating the source of this issue.\par
\begin{figure}[!ht]
%\vspace{-1.9cm}
\centering
\begin{subfigure}[t]{0.9\linewidth}\centering\includegraphics[width=1\linewidth, keepaspectratio]{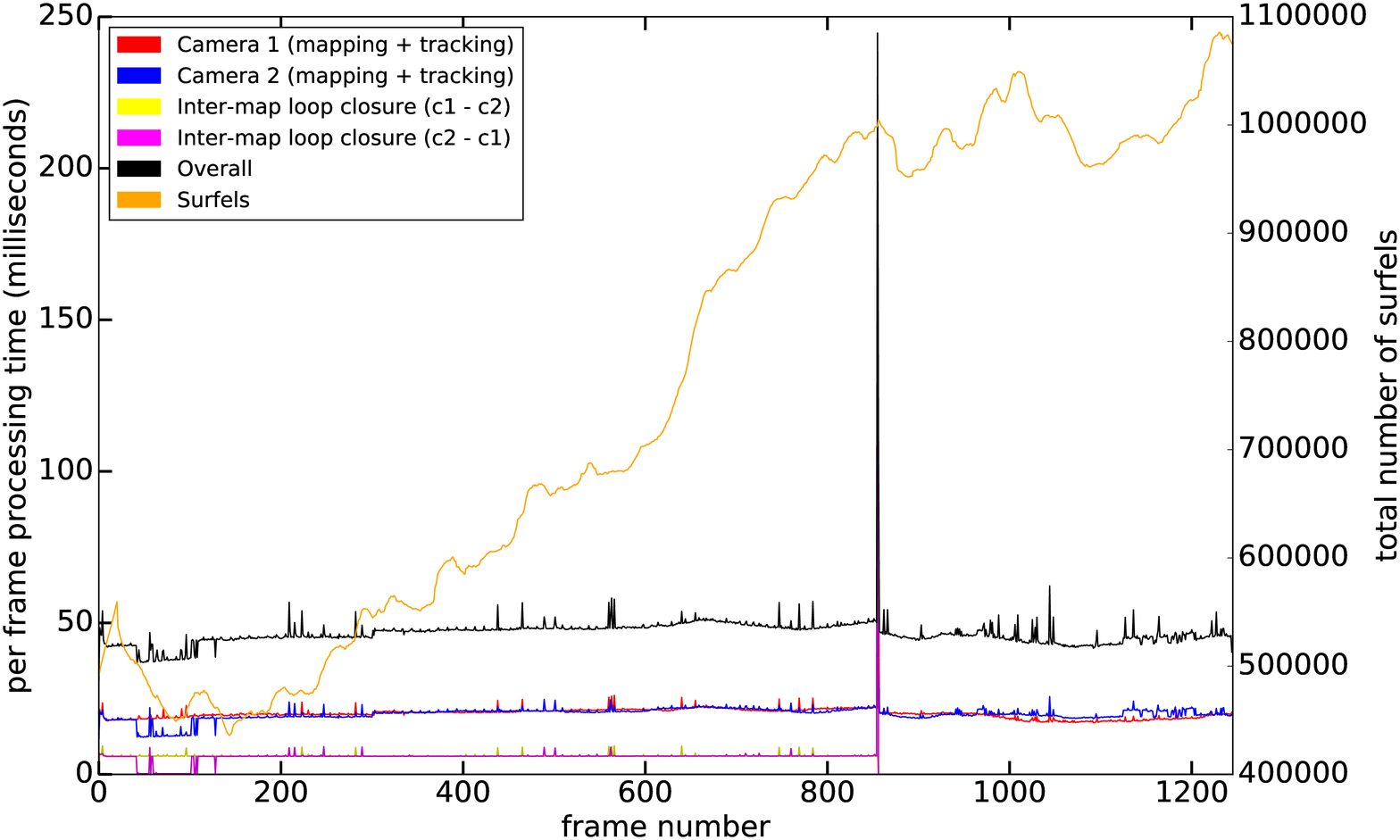}\caption{}\label{timings}\end{subfigure}%0.495

\begin{subfigure}[t]{0.9\linewidth}\centering\includegraphics[width=1\linewidth, keepaspectratio]{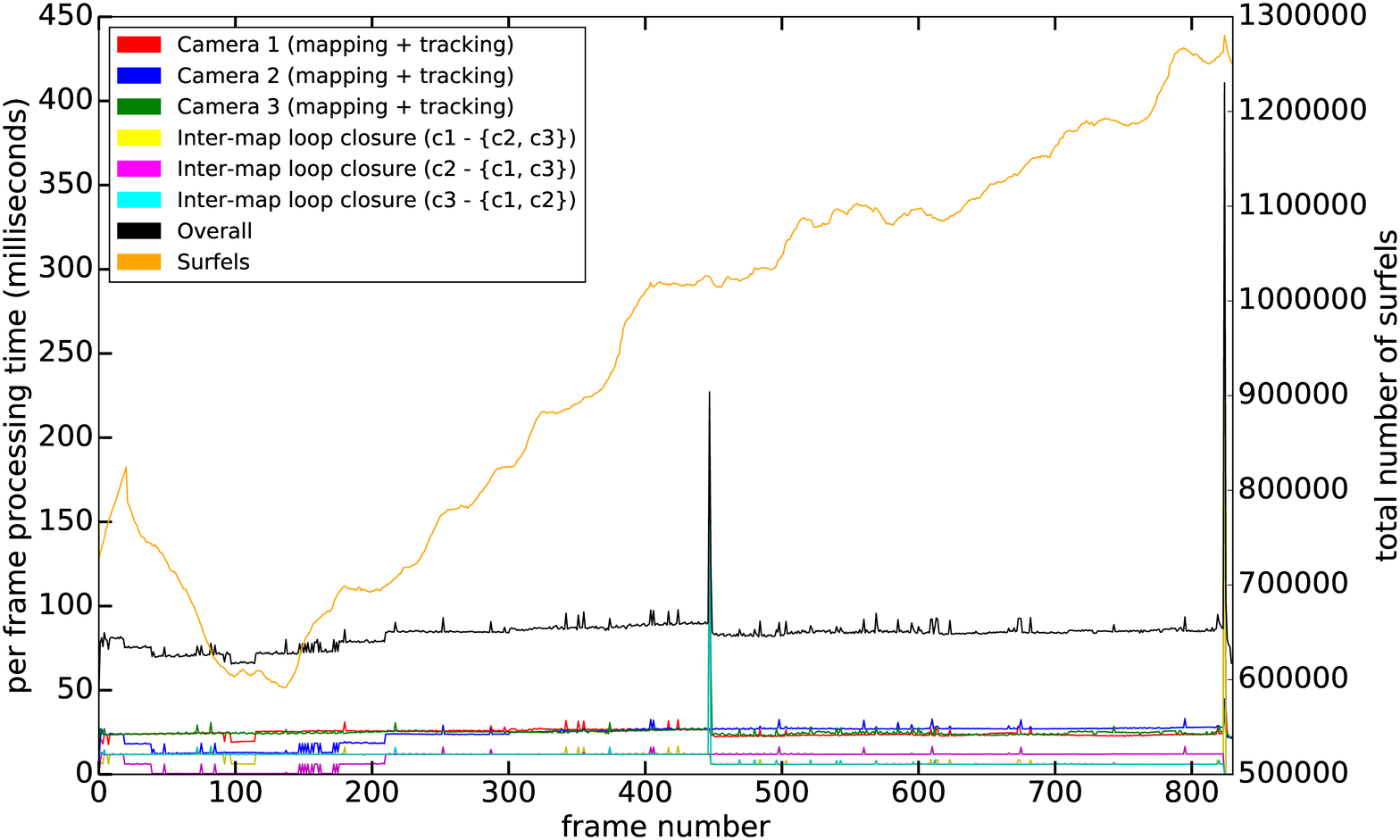}\caption{}\label{timings3}\end{subfigure}
%\vspace{-1.3cm}
\caption[Timings for $2$ camera collaborative mapping session.]{Timings for (\subref{timings}) a $2$ camera and (\subref{timings3}) a $3$ camera collaborative mapping session with the `fr3/office' sequence. The original sequence contained $2487$ frames. For graph (\subref{timings}) we split these frames into two subsequences of equal length, similarly, for graph (\subref{timings3}) we split them into three subsequences of equal length. The subsequences were then processed collaboratively. For example, in (\subref{timings}) Camera $1$ corresponds to the first half of the frames from the original sequence while Camera $2$ corresponds to the second half of the frames.
%In this experiment, camera $1$ represents the first subsequence while camera $2$ represents the second subsequence. 
The spike at frame $942$ in graph (\subref{timings}) is caused by an inter-map loop closure between the two cameras. Similary the two spikes in graph (\subref{timings3}) at frames $449$ and $826$ are both caused by inter-map loop closures.}
\label{fig:timings}
\end{figure}
\begin{figure}[!ht]
\centering
\begin{subfigure}[t]{1.0\linewidth}
\centering
\includegraphics[width=6.cm, height=6.cm, keepaspectratio]{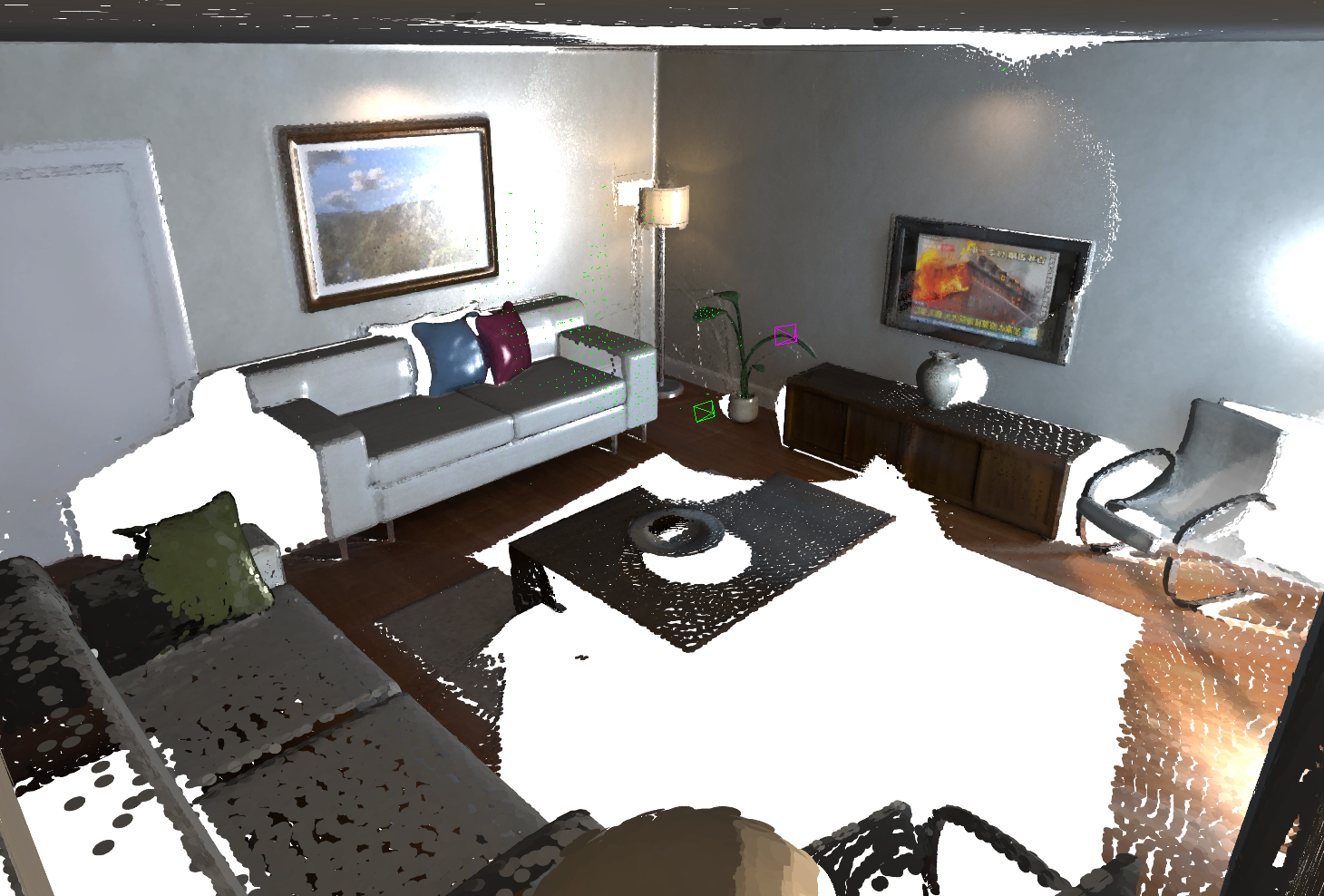}
\hspace{2em}
\includegraphics[width=6cm, height=6.cm, keepaspectratio]{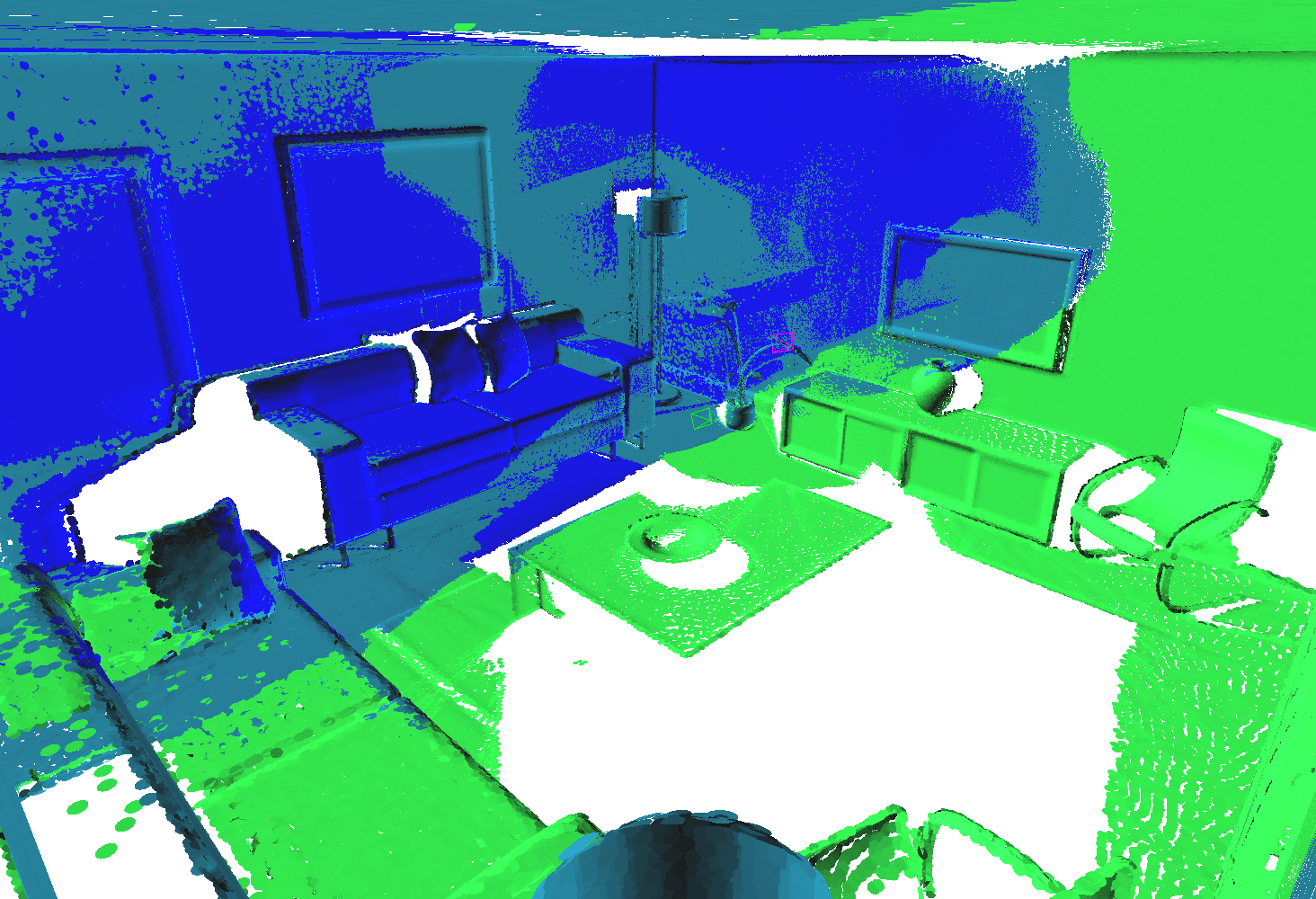}
\caption{}\label{kt0kt1contr}
\end{subfigure}

\begin{subfigure}[t]{1.0\linewidth}
\centering
\includegraphics[width=6.cm, height=6.cm, keepaspectratio]{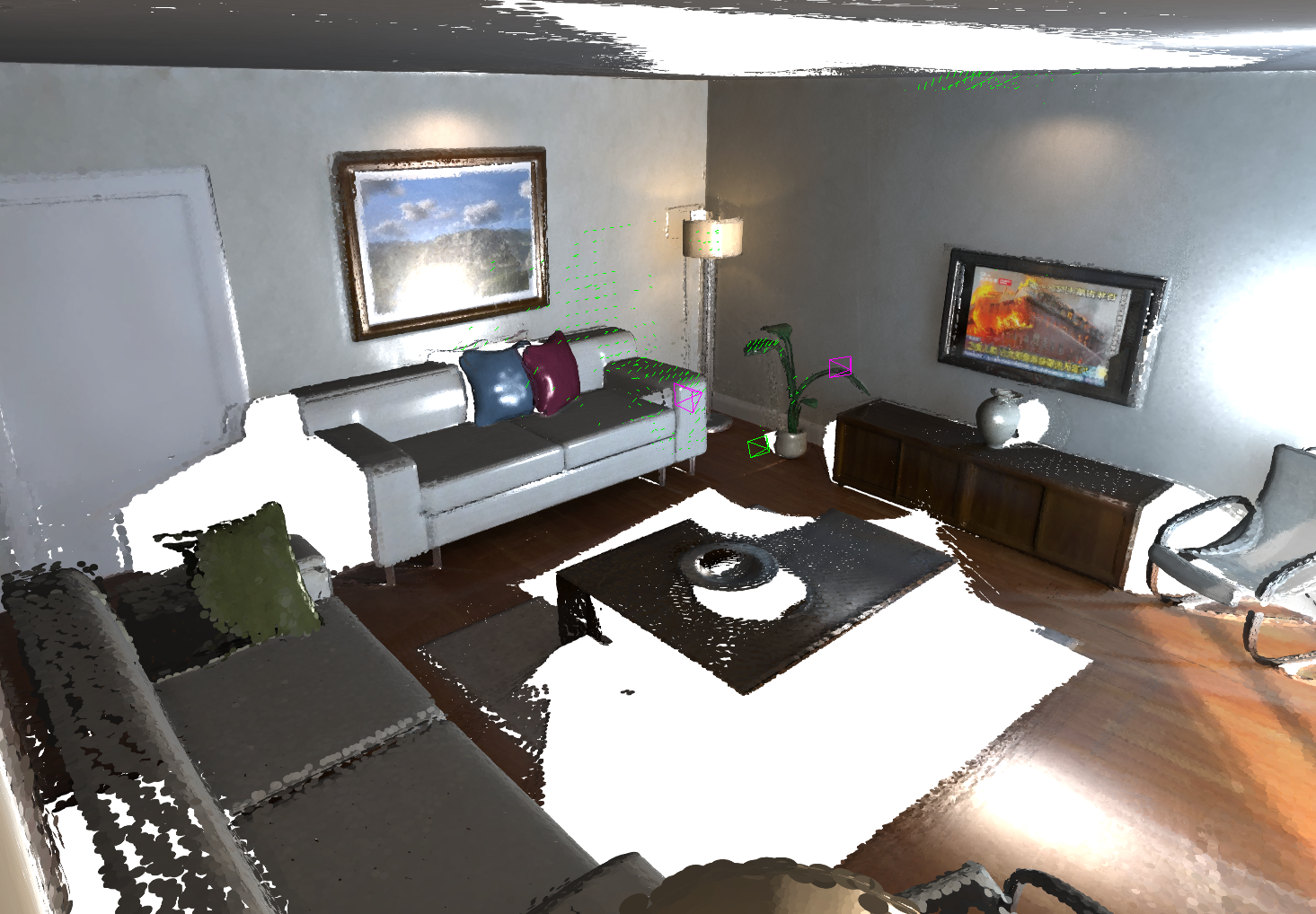}
\hspace{2em}
\includegraphics[width=6.cm, height=6.cm, keepaspectratio]{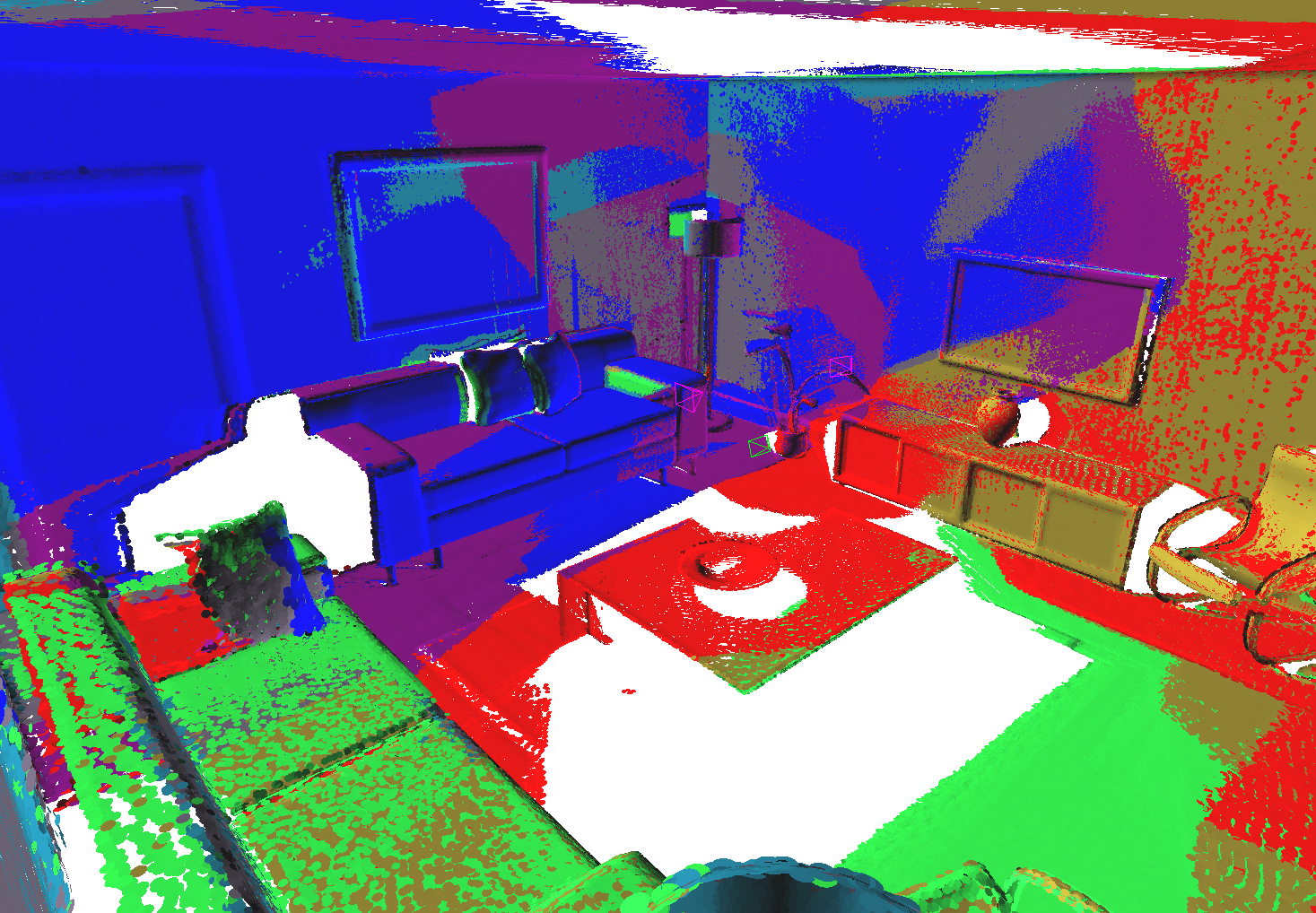}
\caption{}\label{kt0kt1kt2contr}
\end{subfigure}

\begin{subfigure}[t]{1.\linewidth}
\centering
\includegraphics[width=6.cm, keepaspectratio, angle=90,origin=c]{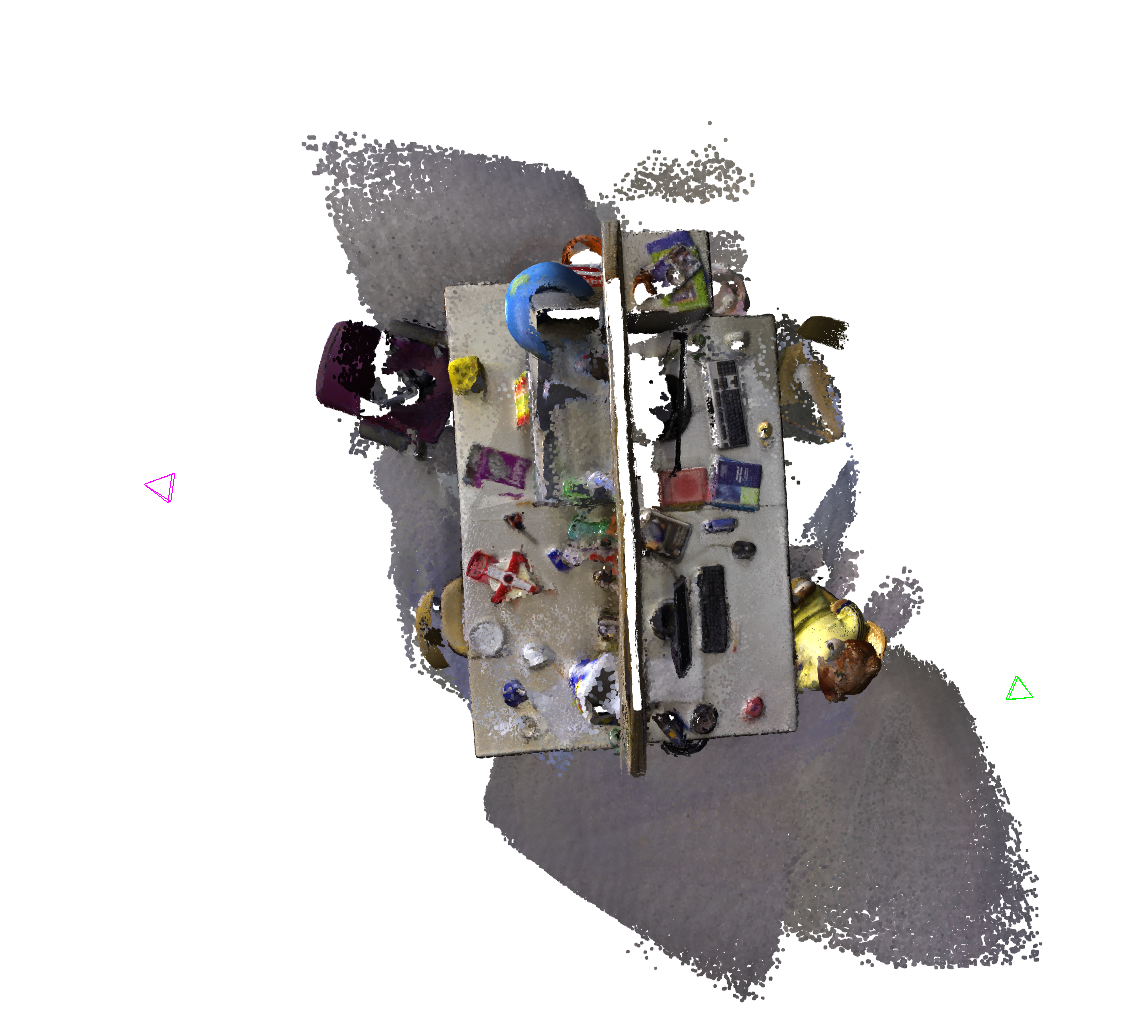}
\hspace{2em}
\includegraphics[width=6cm, keepaspectratio, angle=90,origin=c]{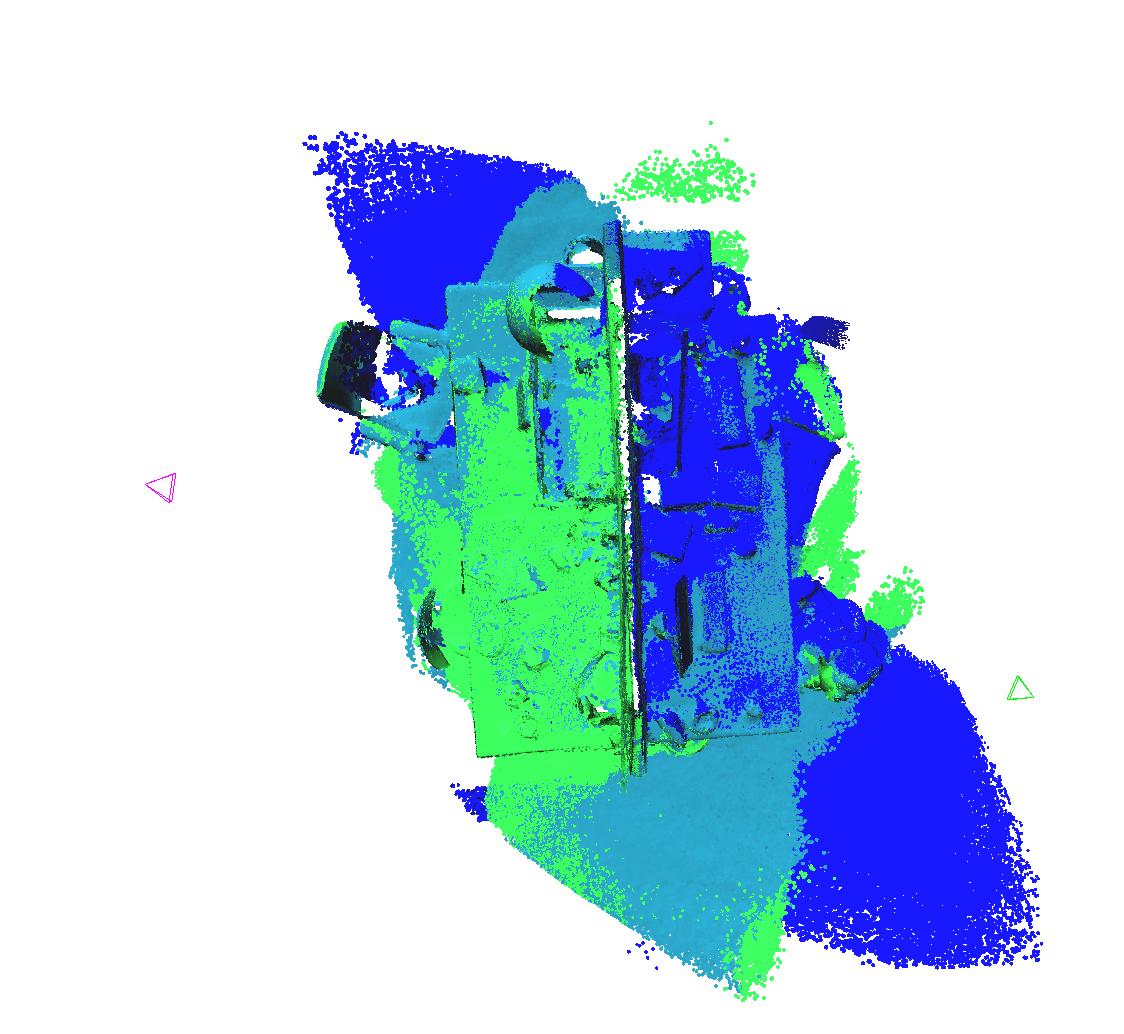}
\caption{}\label{fr3officecontr}
\end{subfigure}

%\begin{subfigure}[t]{0.25\linewidth}\centering\includegraphics[width=4.cm, height=4.cm, keepaspectratio]{images/final_model_kt0kt2_coloured}\label{kt0kt2}\end{subfigure}%\hspace{-2.7cm}
%\begin{subfigure}[t]{0.25\linewidth}\centering\includegraphics[width=4.cm, height=4.cm, keepaspectratio]{images/final_model_kt0kt1kt2_coloured}\label{kt0kt1kt2}\end{subfigure}%\hspace{-2.7cm}
%\begin{subfigure}[t]{0.25\linewidth}\centering\includegraphics[width=5.cm, height=3.cm, keepaspectratio, angle=90,origin=c]{images/fr3_office_col}\label{fr3officecoloured}\end{subfigure}

%\begin{subfigure}[t]{0.25\linewidth}\centering\includegraphics[width=4cm, height=4.cm, keepaspectratio]{images/final_model_kt0kt2_contr}\caption{}\label{kt0kt1contr}\end{subfigure}
%\begin{subfigure}[t]{0.25\linewidth}\centering\includegraphics[width=4.cm, height=4.cm, keepaspectratio]{images/final_model_kt0kt1kt2_contr}\caption{}\label{kt0kt1kt2contr}\end{subfigure}
%\begin{subfigure}[t]{0.25\linewidth}\centering\includegraphics[width=5cm, height=3.cm, keepaspectratio, angle=90,origin=c]{images/fr3_office_con}\caption{}\label{fr3officecontr}\end{subfigure}
\caption[Examples of maps produced by the system.]{Examples of maps produced by the system. The left column shows the final models produced by the system. The right column shows the contributions of each camera to the final model. Each cameras specific contributions are coloured with a separate primary colour. Composite coloured surfels have fused measurements from multiple cameras. Rows (\subref{kt0kt1contr}) and  (\subref{kt0kt1kt2contr}) correspond to the ICL-NUIM $kt_0$ \& $Kt_2$ and $kt_0$ \& $kt_1$ \& $kt_2$ sessions respectively. Row (\subref{fr3officecontr}) corresponds to the $2$ camera session using the `fr3/office' sequence from the Freiburg dataset.}
% The left hand column show the final models produced by the system. The right hand column shows the contributions of each sensor to the final map. The models shown in the top row correspond to the  $kt_0$ \& $kt_2$ session, the ones in the bottom row correspond to the $kt_0$ \& $kt_1$ \& $kt_2$ mapping session.}
\label{fig:models}
\end{figure}

 With the remaining trajectories we looked at the surface reconstruction accuracy and camera trajectory estimation accuracy of the system. To measure the surface reconstruction accuracy we processed an exhaustive set of combinations of the living room trajectories. For each combination we then computed the average per-vertex error in metres using the reference ground truth model. To measure the estimated trajectory accuracy we 
 %also recorded the estimated poses of each camera in each of the combinations. We then 
 computed the average trajectory error root mean square error (ATE RMSE) in metres between each of the estimated trajectories in each of the combinations and the corresponding reference ground truth trajectories. Table \ref{table:2} and its caption lists the specific trajectory combinations we used in these experiments and the results we recorded. Additionally we have provided pose estimation accuray results using the TUM-RGBD datatset of \cite{freiburg}. We used the `fr3/office' sequence, as it is a long and challenging sequence. See Figure \ref{fig:timings} and its caption for details of how we generated collaborative mapping sessions with this dataset.%\par
%To experiment with how well FGR performs at batch aligning submaps we purposefuly reject visual intermap loop closure detections during collaborative mapping to simulate failure cases leading to multiple submaps at the end of a session. We report surface reconstruction accuracy and ATE RMSE for each of the ICL-NUIM and TUM-RGBD trajectory combinations used above. The results are shown in \ref{table:2} alongside the results for fern-based intermap loop closures.\par  
 %The reader will note that we left the fourth sequence, $kt_3$, out of the collaborative experiments. When including this sequence we found that the system experienced poor surface reconstruction. We are currently investigating why exactly this is the case.
The models shown in Figure \ref{fig:models} demonstrate the quality of the maps produced by our system.\par

%We use the synthetic dataset of \cite{iclnuim} to demonstrate the limitations of EFs existing loop closure mechanism for solving inter-map loop closures in our system, across all relevant sequences in the dataset. \cite{iclnuim} contains two scenes, one of a living room and one of an office. There are four sequences per scene. For each scene we push all sequences through our system concurrently. This data is summarised in Table \ref{table:1}.

%\begin{minipage}{0.45\textwidth}
%\scalebox{0.6}{
%\begin{tabular}{|p{2cm} |c | c | c | c | c |c | c | c | c|} 
% \hline
% \text{ } & lr/kt0  & lr/kt1 & lr/kt2 & lr/kt3 & of/kt0  & of/kt1 & of/kt2 & of/kt3\\ [0.5ex] 
% \hline
% inter-map loop closure detections & 0 & 0 & 0 & 0 & 0 & 0 & 0 & 0 \\ \hline
% mean residual error & 0 & 0 & 0 & 0 & 0 & 0 & 0 & 0\\
% \hline
%\end{tabular}}
%\captionof{table}{The top row gives the name of the sequnce and the scene it is from, the next row gives the number of inter-map loop closures detected. The last row gives the per sequence average residual alignment error.}
%\label{table:1}

%\end{minipage}
%\qquad

%\begin{minipage}{0.45\textwidth}

\begin{table}[!ht]
%\vspace{-.5cm}
	\centering
	\scalebox{0.9}{%0.58
	\begin{tabular}{| c | c | c | } 
		\hline
		\multicolumn{3}{| c |}{\textit{Surface reconstruction and trajectory estimation accuracy for both the ICL-NUIM and TUM datasets.}}\\
		\hline	
	%	\multicolumn{1}{| c ||}{}
	%	&
	%	\multicolumn{2}{ c ||}{Fern-based Intermap Loop Closures}
	%	&
	%	\multicolumn{2}{| c |}{FGR-based Batch Alignment}\\
		%\hline
		Sequences & Surface Reconstruction Accuracy & Individual Trajectory ATE RMSE\\
 		\hline
 		%\multicolumn{3}{|c|}{bf{Sequences of One Camera}}\\
% 		\multicolumn{1}{|c||}{} &
 		 \multicolumn{3}{|c|}{ICL-NUIM}\\
 		\hline
 		 %$kt_0$ & $0.0062m$ & $0.0143m$ \\
 		 $kt_0$ & $0.007m$ & $0.014m$ \\ % & $-$ & $-$ \\
		 %$kt_1$ & $0.0069m$ & $0.0127m$ \\
		 $kt_1$ & $0.008m$ & $0.012m$ \\% & $-$& $-$\\
		 %$kt_2$ & $0.0065m$ & $0.0191m$ \\
		 $kt_2$ & $0.009m$ & $0.027m$ \\%& $-$ & $-$\\
		 %$kt_3$ & $0.0308m$ & $0.2219m$ \\
		 $kt_3$ & $0.058m$ & $0.308m$ \\% & $-$ & $-$\\ 
 		\hline
 		%\multicolumn{3}{|c|}{\textbf{Sequences of Two Cameras}}\\
 		\hline
 		%$kt_0$ \& $kt_1$ & $0.0066m$ & $0.0147m$ | $0.5950m$ \\
 		$kt_0$ \& $kt_1$ & $0.007m$ & $0.016m$ | $0.014m$\\% & $0.008m$ & $0.012m$ | $0.014m$\\
 		%$kt_0$ \& $kt_2$ & $0.0066m$ & $0.1112m$ | $0.0191m$\\
 		$kt_0$ \& $kt_2$ & $0.009m$ & $0.016m$ | $0.026m$\\% & $0.009m$ & $0.014m$ | $0.027m$\\
 		%$kt_0kt_3$ & $0.0316$ & $0.0288m$ | $0.1935m$ \\
 		$kt_1$ \& $kt_2$ & $0.01m$ & $0.0124m$ | $ 0.027m$ \\ %& $0.009m$ & $0.017m$ | $0.028m$\\
 		%$kt_1kt_3$ & $0$ & $0$ | $0$ \\
 		%$kt_2kt_3$ & $0$ & $0$ | $0$ \\
 		\hline
 		%\multicolumn{3}{|c|}{\textbf{Sequences of Three Cameras}}\\
 		\hline
 		$kt_0$ \& $kt_1$ \& $kt_2$ & $0.009m$ & $0.015m$ | $0.016m$ | $0.027m$ \\%& $0.009m$ & $0.014m$ | $0.012m$ | $0.027m$\\
 		%$kt_0kt_1kt_3$ & $0$ & $0$ \\
 		%$kt_0kt_2kt_3$ & $0$ & $0$ \\
 		%$kt_1kt_2kt_3$ & $0$ & $0$ \\
 		\hline
% 		\multicolumn{1}{|c||}{} &
 		\multicolumn{3}{|c|}{TUM RGBD}\\
 		\hline
 		$office_0$ & $-$ & $0.018m$ \\%& $-$ & $-$\\
 		\hline
 		\hline
		$office_0$ \& $office_1$ & $-$ & $0.019m$ | $0.015m$\\% & $-$ & $0.014m | 0.015m$\\
		\hline
 		\hline
		$office_0$ \& $office_1$ \& $office_2$ & $-$ & $0.014m$ | $0.015m$ | $0.014m$\\% & $-$ & $0.014 |0.013m | 0.016m$\\
		\hline
% 		\multicolumn{3}{|c|}{\textbf{Sequences of Four Cameras}}\\
% 		\hline
% 		$kt_0kt_1kt_2kt_3$ & $0$ & $0$ \\
% 		\hline
 		
	\end{tabular}
   }
   \vspace{-5pt}
\caption[Surface reconstruction accuracy and trajectory estimation results from the lab dataset in metres.]{Surface reconstruction accuracy and trajectory estimation results from the ICL-NUIM and TUM datasets in metres\footnotemark . The first column in each row gives the specific combination of trajectories used in that session. The second column gives the per collaboration surface reconstruction accuracy. In the last column we give the per sequence ATE RMSE for each collaboration. % The left column shows the specific combinations used in each session, the middle column shows the results achieved using the fern-based intermap loop closures and the right column shows the results achieved using FGR-based batch alignment.  
%The first column in each row gives the specific combination of the four living room trajectories used in that session. The second column gives the per collaboration surface reconstruction accuracy. In the last column we give the per sequence ATE RMSE for each collaboration.
%The reader will note that in the one camera sessions our results are not exactly as reported in \cite{elasticfusionjournal}. This is for a number of reasons; (i) the system is sensitive to the specific GPU of the machine it is deployed on; (ii) as fern encoding is a random process loop closures will not always occur at exactly the same point. %It is also implied that per dataset parameter tuning was performed to improve performance.
 %In addition, while we have made an effort to tune EFs parameters appropriately we have intentionally tried to avoid overfitting, albeit at the cost of decreased performance in certain sequences. Indeed, the approach of parameter tuning becomes impractical as the number of parameters that must be tuned grows in proportion with the number of collaborators in a session.
 }
\label{table:2}
\end{table}

%Certain of the trajectory estimates for the TUM-RGBD dataset are notably higher than as reported in the original EF algorithm. We show that this error is due to the intermap loop closure optimisation. 
Following this, 
%we used the real-world RGB-D dataset of \cite{freiburg}
we again used the `fr3/office' sequence, this time to measure the processing time of the different components of the system during a $2$ camera and then a $3$ camera collaborative session.
% The Freiburg dataset consists of a number of sequences designed to test different aspects of a SLAM systems performance. 
%We used the `Freiburg 3 long office household' sequence, as it is a long and challenging sequence, and split it into $2$ subsequences which we processed collaboratively.
The results of this experiment are shown in Figure \ref{fig:timings}.\par

\footnotetext{The reader will note that in the single camera sessions our results are not exactly as reported in \cite{elasticfusionjournal}. This is for a number of reasons; (i) the system is sensitive to the specific GPU of the machine it is deployed on; (ii) as fern encoding is a random process loop closures will not always occur at exactly the same point. %It is also implied that per dataset parameter tuning was performed to improve performance.
 %In addition, while we have made an effort to tune EFs parameters appropriately we have intentionally tried to avoid overfitting, albeit at the cost of decreased performance in certain sequences. Indeed, the approach of parameter tuning becomes impractical as the number of parameters that must be tuned grows in proportion with the number of collaborators in a session.
 }
%\begin{table}[!ht]
%	\centering
%	\scalebox{1.0}{
%	\begin{tabular}{| c | c | c | c | c |} 
%		\hline
%		Name of sequence & Fr. 1 Room & Fr. 1 Floor & Fr. 2 desk & Fr. 3 office \\
 %		\hline
 %		Number of frames & $1361$ & $1225$ & $2891$ & $2486$ \\
 %		
 %		Number of subsequences & x & x & $2$ & $3$ \\
%
 %		Number of surfels & x & x & $608083$ & $582131$ \\
 %		
 %		Number of ferns harvested & x & x & $1$ & $2$ \\
%
 %		Number of inter-map loop closures & x & x & $1$ & $2$ \\
 %		
 %		Average time to close a loop  & x & x & x & x \\		

 %		\hline
 		
	%\end{tabular}
   %}
   %\vspace{-5pt}
%\caption[Freiburg dataset evaluation results.]{ 
%Freiburg dataset evaluation results.}
%\label{table:3}
%\end{table}

%\subsection{FGR Batch Alignment Results}
%In this section we present results for the FGR batch alignment. To experiment with how well FGR handles EF models we purposefuly reject certain visual intermap loop closure detections during collaborative mapping to simulate failure cases leading to multiple submaps at the end of a session. We look at 3 cases: (i) a two camera session with 0 intermap loop closures (ii) a three camera session with 0 intermap loop closures and (iii) a three camera session with 1 intermap loop closure. In each case we use FGR to batch align all the remaining submaps into a single coordinate frame and measure the surface reconstruction accuracy of the aligned model with respect to the groundtruth. The results are shown in \ref{table:fgr}.\par  

\section{Conclusion}\label{conclusion}

We have described a dense collaborative mapping system capable of reconstructing maps with multiple independently moving cameras, in real-time. ElasticFusion formed the foundation of our system, providing camera tracking and surface fusion capabilities.% We identified two distinct phases of operation during collaborative mapping; an initial phase where cameras maintain submaps independently of one another, and a multi-camera fusion phase where cameras who's local submaps overlap have been merged.
 We used fern-based visual recognition to identify overlap between camera submaps and EF's camera tracking mechanism to find the transformations between the coordinate frames of the matched submaps.\par %The process of matching views and merging maps continues until either processing ends or there is only one global map left.\par
We provided a quantitative analysis of our system using the ICL-NUIM dataset to measure its surface reconstruction accuracy and the camera trajectory estimation accuracy. %We simulated collaborative mapping sessions by processing different combinations of the three living room scene trajectories in parallel. 
%Further to this,
We then used the Freiburg dataset to give a breakdown of frame processing times for the system.\par

Currently the system works well for mapping with $2$ or $3$ cameras, however, mapping with a larger number of cameras would involve overcoming significant challenges. For example, to maintain real-time operation, a more sophisticated approach to mapping and tracking, involving processing each frame in parallel, would be required.
%sequentially tracking each camera and fusing its measurements into the model independently would require no longer reach the throughput requirements to operate in real-time. A more sophisticated approach involving processing each camera in parallel would be required. 
Another direction for future work is the modelling of scene dynamics, a pertinent problem in the context of collaborative mapping as cameras that are reconstructing a scene would inevitably move into and out of view of one another, almost guaranteeing some level of scene dyamics.\par

%EF's global loop closure mechanism breaks down somewhat during multi-camera fusion.
Recall from Section \ref{ef} that the deformation graph influence region for a surfel is computed based on a \textit{temporal-spatial} neighborhood. %criterion.
% This mechanism is invaluable during loop closures as it provides a means of separating source surfaces from destination surfaces, allowing multiple revisitations of the same physical location.
 However, in a collaborative session, cameras which are close in time may be fusing measurements into the model that are far apart in space, meaning that searching the graph for temporally nearby nodes may not yield a spatially nearby influence region  with the knock on effect of inhibiting deformation graph optimisation. Thus the influence region computation needs to be rethought in future work.\par

%\section{Acknowledgments}
%This research was funded by the IRC GOIPG scholarship scheme.%Irish Research Councils Government of Ireland Postgraduate Scholarship scheme. 
\clearpage
\bibliographystyle{apalike}
\bibliography{imvip2018}

\end{document}